\documentclass[10pt,twocolumn,letterpaper]{article}

\usepackage{iccv}
\usepackage{times}
\usepackage{epsfig}
\usepackage{graphicx}
\usepackage{amsmath}
\usepackage{amssymb}

\usepackage[pagebackref=true,breaklinks=true,letterpaper=true,colorlinks,bookmarks=false]{hyperref}

\usepackage{caption}
\usepackage{adjustbox}
\usepackage{wrapfig}
\usepackage{multirow}

\usepackage{algorithm}
\usepackage{algpseudocode}
\usepackage{kotex}
\usepackage[numbers,sort]{natbib}
\usepackage{verbatim}
\usepackage{lipsum}
\usepackage[accsupp]{axessibility} 

\newcommand\blfootnote[1]{%
  \begingroup
  \renewcommand\thefootnote{}\footnote{#1}%
  \addtocounter{footnote}{-1}%
  \endgroup
}

\iccvfinalcopy 

\def\iccvPaperID{2958} 

\ificcvfinal\fi

\begin{document}

\title{ILVR: Conditioning Method for Denoising Diffusion Probabilistic Models}

\author{Jooyoung Choi$^1$ ~~~~~~~ Sungwon Kim$^1$ ~~~~~~~ Yonghyun Jeong$^3$  ~~~~~~~ Youngjune Gwon$^3$ ~~~~~~~ Sungroh Yoon$^{1, 2, }$\thanks{}\\
$^1$ Data Science and AI Laboratory, Seoul National University, Korea\\
$^2$ ASRI, INMC, and Interdisciplinary Program in AI, Seoul National University, Korea\\
$^3$ AI Research Team, Samsung SDS\\
{\tt\small \{jy\_choi, ksw0306, sryoon\}@snu.ac.kr ~~~~~~~ \{yhyun.jeong, gyj.gwon\}@samsung.com}}

\twocolumn[{
\renewcommand\twocolumn[1][]{#1}
\vspace{-4.4em}
\maketitle
\vspace{-2.5em}
\begin{center}
\centering
        \centering
        \includegraphics[width=0.93\linewidth]{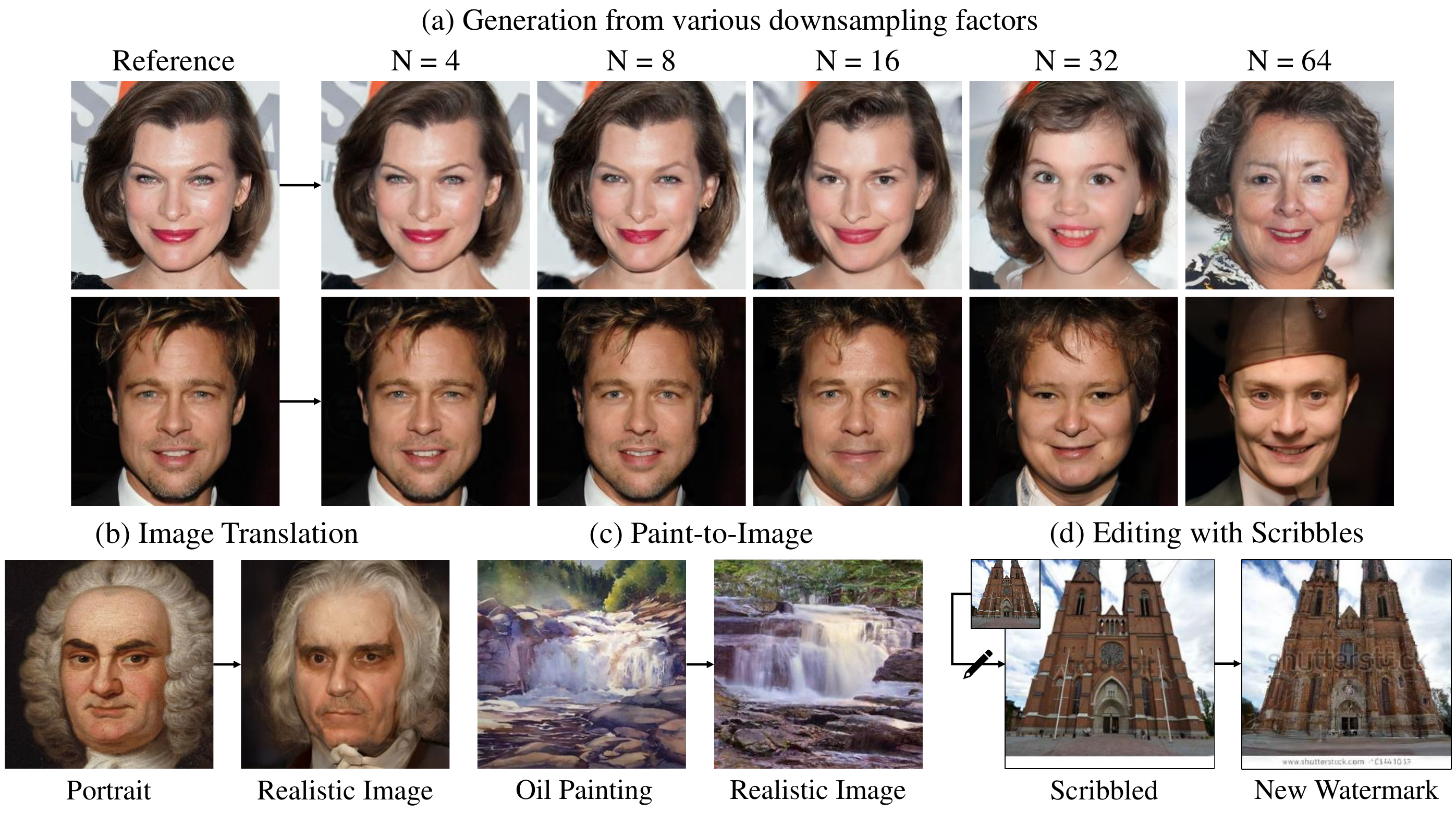}
        \captionof{figure}{\textbf{Iterative Latent Variable Refinement for DDPM.} Our method of controlling Denoising Diffusion Probabilistic Model (DDPM) motivates various image generation tasks such as: (a) Generating from various downsampling factors; (b) Image translation; (c) Paint-to-image; and (d) Editing with scribbles.}
        \vspace{0.7em}
        \label{fig:teaser}
\end{center}
}]

\ificcvfinal\thispagestyle{empty}\fi


\begin{abstract}
\blfootnote{$*$Correspondence to: Sungroh Yoon (sryoon@snu.ac.kr)}
Denoising diffusion probabilistic models (DDPM) have shown remarkable performance in unconditional image generation. However, due to the stochasticity of the generative process in DDPM, it is challenging to generate images with the desired semantics. In this work, we propose Iterative Latent Variable Refinement (ILVR), a method to guide the generative process in DDPM to generate high-quality images based on a given reference image. Here, the refinement of the generative process in DDPM enables a single DDPM to sample images from various sets directed by the reference image. The proposed ILVR method generates high-quality images while controlling the generation. The controllability of our method allows adaptation of a single DDPM without any additional learning in various image generation tasks, such as generation from various downsampling factors, multi-domain image translation, paint-to-image, and editing with scribbles. Our code is available at: \url{https://github.com/jychoi118/ilvr_adm}.
\vspace{-1em}
\end{abstract}

\section{Introduction}

Generative models, such as generative adversarial networks (GAN)~\cite{goodfellow2014generative, stylegan, brock2018large}, normalizing flows \cite{kingma2018glow}, and variational autoencoders \cite{vahdat2020nvae},
have shown remarkable quality in image generation, and have been applied to numerous purposes such as image-to-image translation \cite{park2019semantic, choi2020stargan, shaham2019singan, wang2018high, park2020swapping, gu2020image} and image editing \cite{abdal2019image2stylegan, shen2020interpreting, harkonen2020ganspace}.

There are mainly two approaches to control generative models to generate images as desired: 
one is by designing the conditional generative models for the desired purpose, and the other is by leveraging well-performed unconditional generative models.

The first approach learns to control by providing the desired condition in training procedure and has shown remarkable performance on various tasks, such as segmentation mask conditioned generation \cite{park2019semantic,zhu2020sean}, style transfer~\cite{gatys2016image,yoo2019photorealistic}, and inpainting \cite{yu2019free,liu2018image}.
The second approach utilizes high-quality generative models, such as StyleGAN \cite{stylegan,karras2020analyzing} or BigGAN \cite{brock2018large}. 
Shen \etal \cite{shen2020interpreting} and H{\"a}rk{\"o}nen \etal \cite{harkonen2020ganspace} manipulate semantic attributes of images by analyzing latent space of pre-trained generative models, while Huh \etal \cite{huh2020transforming} and Zhu \etal \cite{zhu2020domain} perform image editing by projecting image into the latent space. 

Denoising diffusion probabilistic models (DDPM)~\cite{ho2020denoising,sohl2015deep}, an iterative generative model, has shown comparable performance to the state-of-the-art models in unconditional image generation.
DDPM learns to model the Markov transition from simple distribution to data distribution and generates diverse samples through sequential stochastic transitions. Samples obtained from the DDPM depend on the initial state of the simple distribution and each transition. 
However, it is challenging to control DDPM to generate images with desired semantics, since the stochasticity of transitions generates images with inconsistent high-level semantics, even from the same initial state.

In this work, we propose a learning-free method, iterative latent variable refinement (ILVR), to condition the generation process in well-performing unconditional DDPM.
Each transition in the generation process is refined utilizing a given reference image. 
By matching each latent variable, ILVR ensures the given condition in each transition thus enables sampling from a conditional distribution. Thus, ILVR generates high-quality images sharing desired semantics. 

We describe user controllability of our method, which enables control on semantic similarity of generated images to the refenence. Fig.~\ref{fig:teaser}(a) and Fig.~\ref{fig:hierarchy} show samples sharing semantics ranging from coarse to fine information.
Besides, reference images can be selected from unseen data domains.
From these properties, we were motivated to leverage unconditional DDPM learned on single data domain to multi-domain image translation; a challenging task where existing works had to learn on multiple data domains.
Furthermore, we extend our method to paint-to-image and editing with scribbles (Fig. ~\ref{fig:teaser}(c) and (d)). We demonstrate that our ILVR enables leveraging a single unconditional DDPM model on these various tasks without any additional learning or models.
Measuring Fréchet Inception Distance (FID) and Learned Perceptual Image Patch Similarity (LPIPS), we confirm that our generation method from various downsampling factors provides control over diversity while maintaining visual quality.

Our paper makes the following contributions:

\begin{itemize}

\item We propose ILVR, a method of refining each transition in the generative process by matching each latent variable with given reference image.

\item We investigate several properties that allows user controllability on semantic similarity to the reference.

\item We demonstrate that our ILVR enables leveraging unconditional DDPM in various image generation tasks including multi-domain image translation, paint-to-image, and editing with scribbles.
\end{itemize}

\section{Background}

Denoising diffusion probabilistic models (DDPM)~\cite{ho2020denoising,sohl2015deep} is a class of generative models that show superior performance~\cite{ho2020denoising} in unconditional image generation. It learns a Markov Chain which gradually converts a simple distribution such as isotropic Gaussian, into a data distribution.
Generative process learns the reverse of the DDPM's forward (diffusion) process, a fixed Markov Chain that gradually adds noise to data when sequentially sampling latent variables $x_{1},...,x_{T}$ of the same dimensionality.
Here, each step in the forward process is a Gaussian translation.
\begin{equation}\label{eq:forward}
q(x_{t}|x_{t-1}):=N(x_{t};\sqrt{1-\beta _{t}}x_{t-1},\beta _{t}\mathbf{I}),
\end{equation}
where $\beta_{1},...,\beta_{T}$ is a fixed variance schedule rather than learned parameters~\cite{ho2020denoising}. 
Eq.~\ref{eq:forward} is a process finding $x_{t}$ by adding a small Gaussian noise to the latent variable.
Given clean data $x_{0}$, sampling of $x_{t}$ is expressed in a closed form:
\begin{equation}\label{eq:property_derivation}
q(x_{t}|x_{0}):=N(x_{t};\sqrt{\overline{\alpha}_{t}}x_{0},(1-\overline{\alpha}_{t})\mathbf{I}),
\end{equation}
where $\alpha _{t}:=1-\beta _{t}$ and $\overline{\alpha}_{t}:=\prod_{s=1}^t \alpha _{s}$. 
Therefore, $x_{t}$ can be expressed as a linear combination of $x_{0}$ and $\epsilon$:
\begin{equation}\label{eq:property}
x_{t}=\sqrt[]{\overline{\alpha} _{t}}x_{0}+\sqrt[]{1-\overline{\alpha} _{t}}\epsilon,
\end{equation}
where $\epsilon \sim N(0,\mathbf{I})$ has the same dimensionality as data $x_{0}$ and latent variables $x_{1},...,x_{T}$.

Since the reverse of the forward process $q(x_{t-1}|x_{t})$ is intractable, DDPM learns parameterized Gaussian transitions $p_{\theta}(x_{t-1}|x_{t})$.
The generative (or reverse) process has the same functional form~\cite{sohl2015deep} as the forward process, and it is expressed as a Gaussian transition with learned mean and fixed variance~\cite{ho2020denoising}:
\begin{equation}\label{eq:reverse}
p_{\theta}(x_{t-1}|x_{t})=N(x_{t-1};\mu_{\theta}(x_{t},t),{\sigma}_t^2\mathbf{I}).
\end{equation}

\begin{figure*}[t!]
    \centering
    \includegraphics[width=0.88\linewidth]{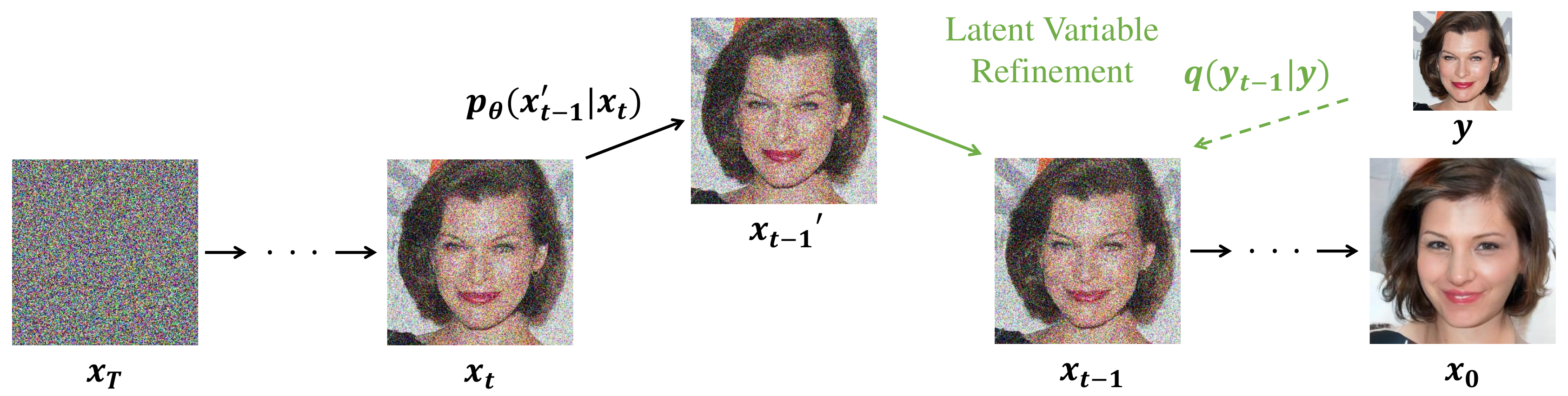}
    \caption{\textbf{Graphical model of Iterative Latent Variable Refinement.} From state $x_{t}$, we first sample unconditional proposal $x_{t-1}$ according to Eq.~\ref{eq:sampling}. Then, we match latent variable with encoded condition $y_{t-1}$ according to Eq.~\ref{eq:match}.}
    \label{fig:graph}
\end{figure*}

Further, by decomposing $\mu_{\theta}$ into a linear combination of $x_{t}$ and the noise approximator $\epsilon_{\theta}$, the generative process is expressed as:
\begin{equation}\label{eq:sampling}
x_{t-1}=\frac{1}{\sqrt[]{\alpha _{t}}}(x_{t}-\frac{1-\alpha _{t}}{\sqrt[]{1-\overline{\alpha }_{t}}}\epsilon _{\theta }(x_{t},t))+\sigma _{t}\mathbf{z},
\end{equation}

where $\mathbf{z}\sim N(0,\mathbf{I})$, which suggests that each generation step is stochastic. Multiple stochastic process steps result in a difficulty in controlling the DDPM generative process. $\epsilon_{\theta}$ represents a neural network with the same input and output dimensions and the noise predicted by the neural network $\epsilon_{\theta}$ in each step is used for the denoising process in Eq.~\ref{eq:sampling}.

\section{Method}

Leveraging the capabilities of DDPM, we propose a method of controlling unconditional DDPM.
We introduce our method, Iterative Latent Variable Refinement (ILVR), in Section.~\ref{sec:vanilla_method}. Section. \ref{sec:3.2} investigates several properties of ILVR, which motivate control of two factors: downsampling factors and conditioning range.

\subsection{Iterative Latent Variable Refinement}\label{sec:vanilla_method}

In this section, we introduce Iterative Latent Variable Refinement (ILVR), a method of conditioning the generative process of the unconditional DDPM model to generate images that share high-level semantics from given reference images. 
For this purpose, we sample images from the conditional distribution $p(x_0|c)$ with the condition $c$:
\begin{align}\label{eq:jointgivenc}
\begin{split}
p_{\theta}(x_{0}|&c)=\int_{}^{}p_{\theta}(x_{0:T}|c)dx_{1:T}, \\
p_{\theta}(x_{0:T}|c)&=p(x_{T})\prod_{t=1}^T p_{\theta}(x_{t-1}|x_{t},c).
\end{split}
\end{align}

Each transition $p_{\theta}(x_{t-1}|x_t,c)$ of the generative process depends on the condition $c$.
However, the unconditionally trained DDPM represents unconditional transition $p_{\theta}(x_{t-1}|x_t)$ of Eq.~\ref{eq:reverse}. Our ILVR provides condition $c$ to unconditional transition $p_{\theta}(x_{t-1}|x_t)$ without additional learning or models. Specifically, we refine each unconditional transition with a downsampled reference image.

\newcommand{\factorial}{\ensuremath{\mbox{\sc Factorial}}}
\begin{algorithm}[t!]
\caption{Iterative Latent Variable Refinement}\label{alg1}
\begin{algorithmic}[1]
    \State \textbf{Input}: Reference image $y$
    \State \textbf{Output}: Generated image $x$ 
    \State $\phi _{N}(\cdot)$: low-pass filter with scale N
   \State Sample $x_{T}\sim  N(\mathbf{0},\mathbf{I})$
   \For{$t=T,...,1$}
      \State $\mathbf{z}\sim  N(\mathbf{0},\mathbf{I})$ 
      \State $x_{t-1}'\sim p_{\theta}(x_{t-1}'|x_{t})$ \Comment{unconditional proposal}
      \State $y_{t-1}\sim q(y_{t-1}|y)$ \Comment{condition encoding}
      \State $x_{t-1}\leftarrow \phi _{N}(y_{t-1})+x_{t-1}'-\phi _{N}(x_{t-1}')$ 
      
   \EndFor
   \State \textbf{return} $x_{0}$
\end{algorithmic}
\end{algorithm}

Let $\phi _{N}(\cdot)$ denote a linear low-pass filtering operation, a sequence of downsampling and upsampling by a factor of $N$, therefore maintaining dimensionality of the image.
Given a reference image $y$, the condition $c$ is to ensure the downsampled image $\phi_{N}(x_{0})$ of the generated image $x_0$ to be equal to $\phi_{N}(y)$. 

Utilizing the forward process $q(x_t|x_0)$ of Eq.~\ref{eq:property} and the linear property of $\phi _{N}$, each Markov transition under the condition $c$ is approximated as follows: 
\begin{align}\label{eq:condreverse}
p_{\theta}(x_{t-1}|x_{t},~c) \approx p_{\theta}(x_{t-1}|x_{t},\phi _{N}(x_{t-1})=\phi _{N}(y_{t-1})),
\end{align}
where $y_{t}$ can be sampled following Eq.~\ref{eq:property}. The condition $c$ in each transition from $x_t$ to $x_{t-1}$ can be replaced with a local condition, wherein latent variable $x_{t-1}$ and corrupted reference $y_{t-1}$ share low-frequency contents.
To ensure the local condition in each transition, we first use DDPM to compute the unconditional proposal distribution of $x_{t-1}^{'}$ from $x_t$. Then, since operation $\phi$ maintains dimensionality, we refine the proposal distribution by matching $\phi(x_{t-1}^{'})$ of the proposal $x_{t-1}^{'}$ with that of $y_{t-1}$ as follows:
\begin{align}\label{eq:match}
\begin{split}
& x_{t-1}^{'} \sim   p_{\theta}(x_{t-1}^{'}|x_{t}), \\ 
x_{t-1} = ~&\phi(y_{t-1}) + (I-\phi)(x_{t-1}^{'}).
\end{split}
\end{align}

\begin{figure}[t!]
    \centering
    \includegraphics[width=1.0\linewidth]{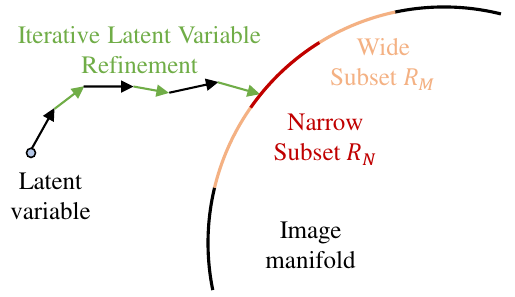}
    \caption{Guiding generation process toward reference directed subset, where $N\leq M$.}
    \label{fig:manifold}
\end{figure}

\begin{figure*}[t!]
    \centering
    \includegraphics[width=1.0\linewidth]{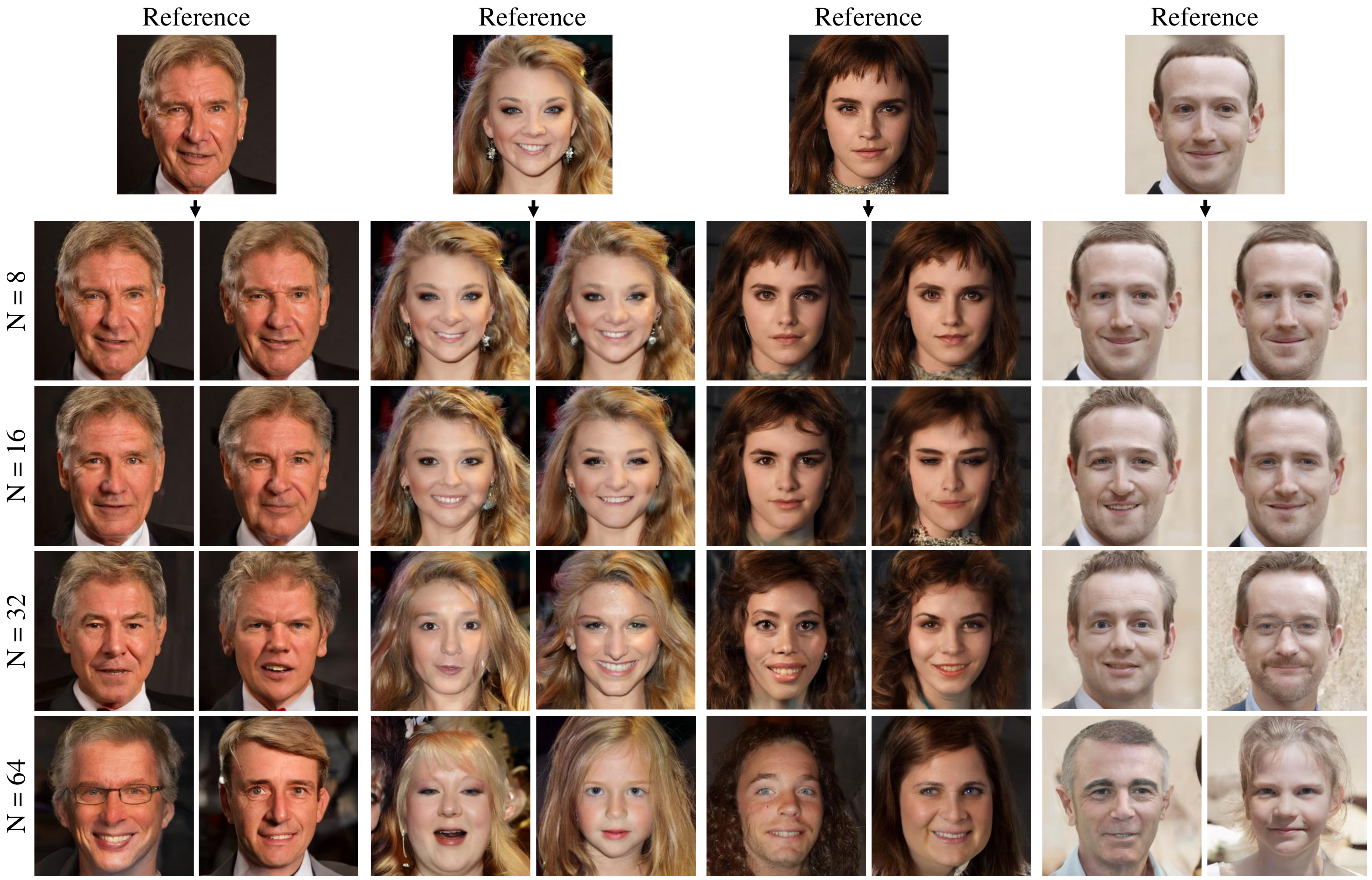}
    \caption{\textbf{Generating from various downsampling factors.} Samples obtained from high factor (N=64) are diverse, only bringing coarse information (color scheme) from reference images. Samples from middle factor (N=32) brought middle level semantics (facial features) while samples from low factors (N=16,8) are highly similar to the reference.}
    \label{fig:hierarchy}
\end{figure*}

By matching latent variables following Eq.~\ref{eq:match}, ILVR ensures local condition in Eq.~\ref{eq:condreverse}, thus enables conditional generation with unconditional DDPM. 
Fig.~\ref{fig:graph} and Algorithm~\ref{alg1} illustrate our ILVR. 
Although we approximate the conditional transition with a simple modification of the unconditional proposal distribution, Fig.~\ref{fig:teaser}(a) and Fig.~\ref{fig:hierarchy} show diverse, high-quality samples sharing semantics of the references.

\begin{figure*}
    \centering
    \includegraphics[width=0.99\linewidth]{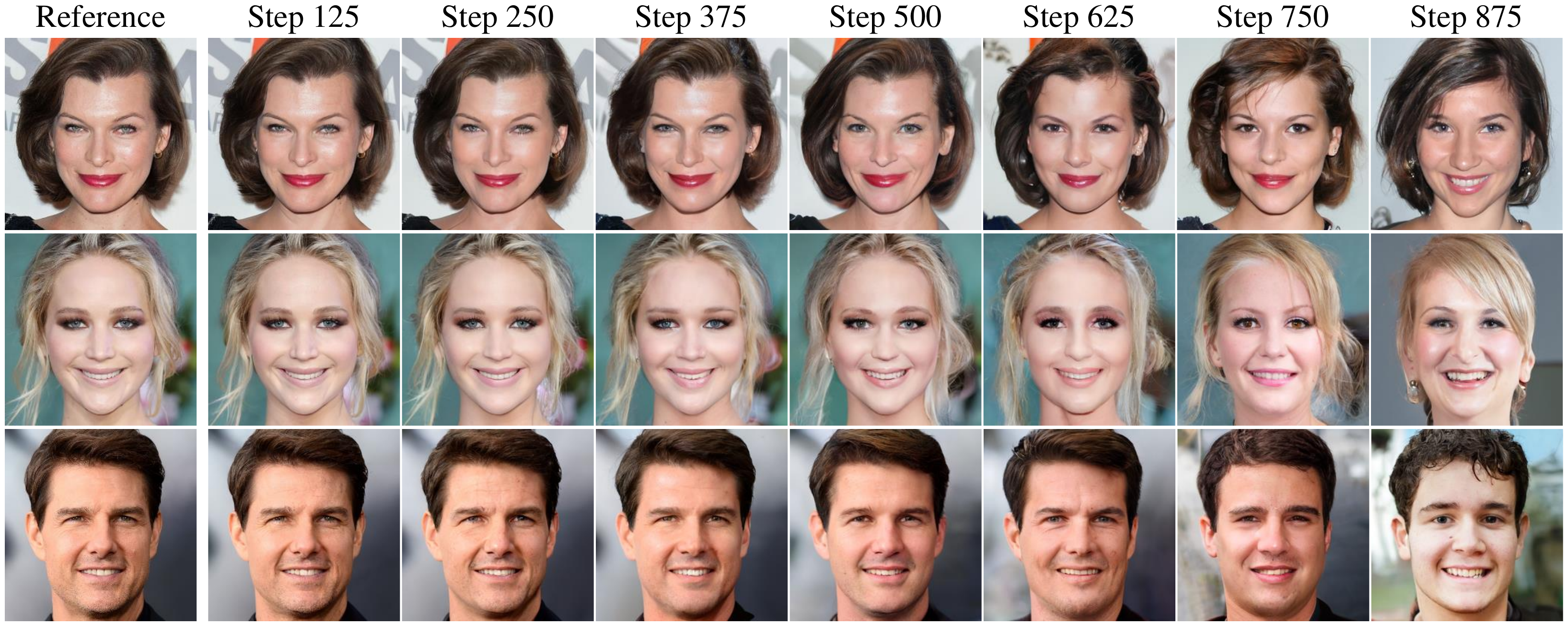}
    \caption{\textbf{Generating from various conditioning ranges.} Samples are generated with ILVR on steps from 1000 to 125, from 1000 to 250, and so on. Samples start to deviate from the reference images with range narrower than step 1000-500.}
    \label{fig:thierarchy}
\end{figure*}

\subsection{Reference selection and user controllability}\label{sec:3.2}

Let $\mu$ be the set of images that an unconditional DDPM can generate.
Our method enables sampling from a conditional distribution with a given reference image $y$. In other words, we sample images from a subset of $\mu$, which is directed by the reference image. 

To extend our method to various applications, we investigate 1) minimum requirement on reference image selection and 2) user controllability on reference directed subset, which defines semantic similarity to the reference.
To provide an intuition for reference selection and control, we investigate several properties. 
Fig.~\ref{fig:manifold} visualizes ILVR in each generation step to guide toward the subset directed by the reference. 

We denote the directed subset as:
\begin{equation}\label{eq:}
R_{N}(y)=\left\{ x : \, \phi_{N}(x)=\phi_{N}(y) \right\},
\end{equation}
representing the set of images $x \in \mathbb{R}^{H\times H}$ which are equivalent to the downsampled reference image $y$.

We consider a range of conditioning steps by extending the above notation:
\begin{equation}\label{eq:nab}
R_{N,~(a,~b)}(y)=\left\{ x : \, \phi_{N}(x_{t})=\phi_{N}(y_{t}), \, t\in[b,~a] \right\},
\end{equation}
where 
$R_{N,~(a,~b)}(y)$ represents the distribution of images matching latent variables (line 9 of Alg.~\ref{alg1}) in steps $b$ to $a$.
We will now discuss several properties on the reference selection and subset control. \\

\noindent\textbf{Property 1}.\hspace{1.0em} Reference image can be any image selected from the set:
\begin{equation}\label{prop:1}
Y=\left\{ y  \, : \, \phi_{N}(y)=\phi_{N}(x), \, x\in \mu\right\},
\end{equation}
the reference image only needs to match the low-resolution space of learned data distribution. Even reference images from unseen data domains are possible. Thus, we can select a reference from unseen data domains and perform multi-domain image translation, as demonstrated in Section.~\ref{sec:domainshift}.
\\

\noindent\textbf{Property 2}.\hspace{1.0em} Considering downsampling factors $N$ and $M$ where $N\leq M$, 
\begin{equation}\label{prop:2}
R_{N}\subset R_{M} \subset \mu,
\end{equation}
which suggests that higher factors correspond to broader image subsets.

As higher factor $N$ enables sampling from broader set of images, sampled images are more diverse and exhibit lower semantic similarity to the reference.
In Fig.~\ref{fig:hierarchy}, perceptual similarity to the reference image is controlled by the downsampling factors.
Samples obtained from higher factor $N$ share coarse features of the reference, while samples from lower $N$ share also finer features. Note that since $R_N$ is a subset of $\mu$, our sampling method maintains the sample quality of unconditional DDPM.
\\

\noindent\textbf{Property 3}.\hspace{1.0em} Limiting the range of conditioning steps enables sampling from a broader subset, while sampling from learned image distribution is still guaranteed.
\begin{equation}\label{prop:3}
R_{N}\subset R_{N,(T,k)} \subset \mu.
\end{equation}

Fig.~\ref{fig:thierarchy} shows the tendency of generated images when gradually limiting the range of conditioned steps. Compared to changing downsampling factors, changing conditioning range has a fine-grained influence on sample diversity. 

\section{Experiments and Applications}\label{sec:experiment}

\begin{figure*}[t!]{}
    \centering
    \includegraphics[width=0.99\linewidth]{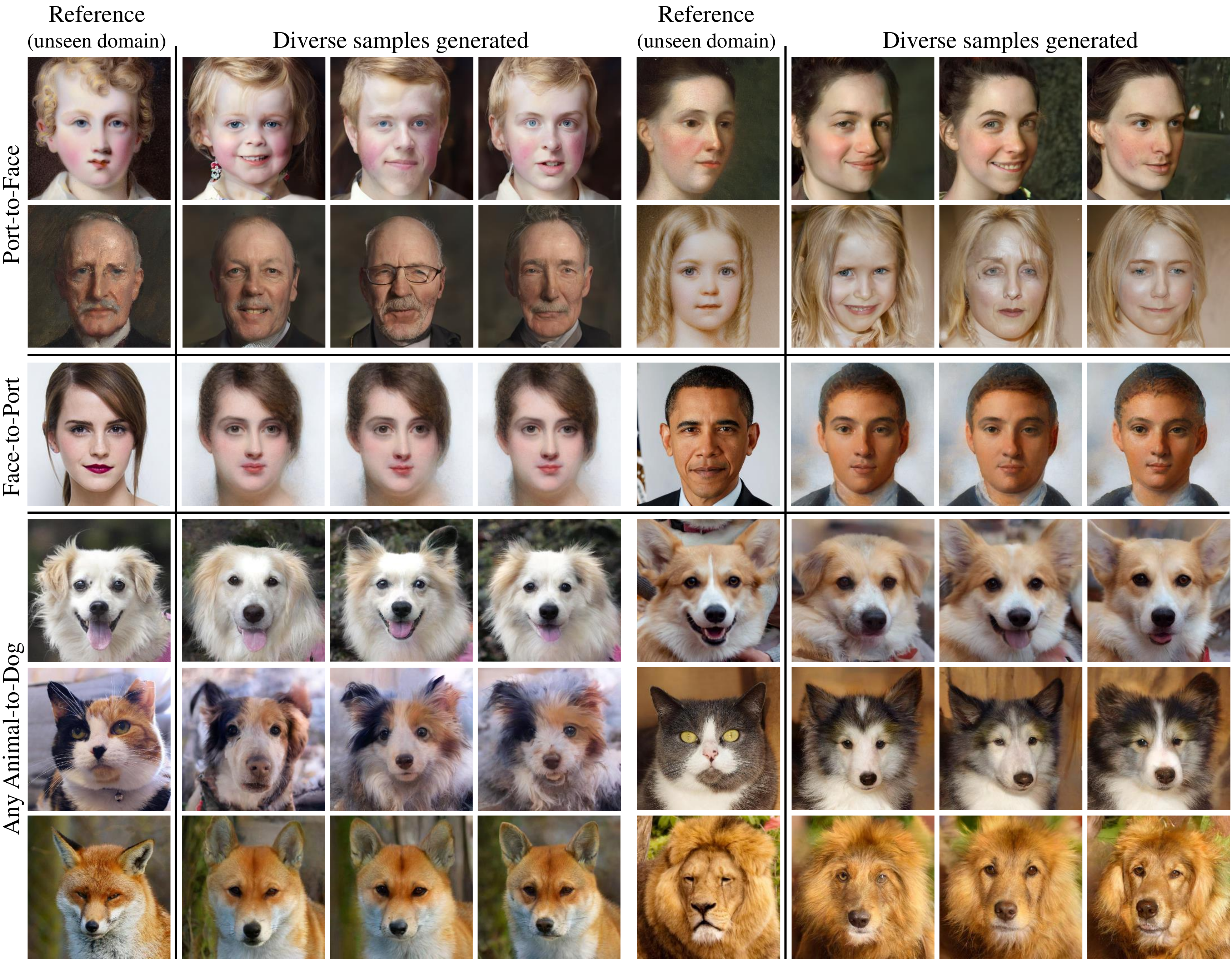}
    \caption{\textbf{Image translation from various source domains.} Row1-2: portrait to face. Row3: face to portrait. Row4-6: any animal (dog, cat, wildlife) to dog. Our method enables translation from any source domains, unseen in the training phase. Moreover, our method generate diverse samples.}
    \label{fig:domainshift}
\end{figure*}

As discussed previously, ILVR generates high-quality images and allows control on semantic similarity to the reference.  We first show qualitative results of controlled generation in Section.~\ref{sec:various_level}. Then we demonstrate ILVR on various image generation tasks in Sections~\ref{sec:domainshift},~\ref{sec:paint}, and~\ref{sec:scribble}. Quantitative evaluations on the visual quality and diversity of ILVR are presented in Section.~\ref{sec:evaluation}.

We trained the DDPM model on FFHQ~\cite{stylegan}, MetFaces~\cite{karras2020training}, AFHQ~\cite{choi2020stargan}, LSUN-Church~\cite{yu2015lsun}, and Places365 datasets~\cite{zhou2017places}, to exemplify its applicability in various tasks. We used correctly implemented resizing library~\cite{ResizeRight} for the operation $\phi_N$.
Reference face images are from the web, those unseen during training. See supplementary materials for details on implementation and evaluations.

\subsection{Qualitative Results on User Controllability}\label{sec:various_level}

Semantic similarity to the reference vary based on the downsampling factor $N$ and the conditioning step range $[b,~a]$.
In Fig.~\ref{fig:hierarchy}, images are generated from the reference image downsampled by various factors. As the factor N increase, samples are more diverse and perceptually less similar to the reference, as stated in Eq.~\ref{prop:2}. For example, samples obtained from N=8 differ with references in fine details (e.g., hair curls, eye color, earring) while samples from N=64 share only coarse features (e.g., color scheme) with the reference. This user controllability on similarity to the reference supports learning-free adaptation of a single pre-trained model to various tasks, as described subsequently. 

In addition to models we reproduced, we also utilize publicly available guided-diffusion~\cite{dhariwal2021diffusion}, recent state-of-the-art DDPM. Fig.~\ref{fig:gddpm} shows samples generated with unconditional models trained on LSUN~\cite{yu2015lsun} datasets. Samples share either coarse (N=64) or fine (N=16) features from the references. Such results suggest that our method can be applied to any unconditional DDPMs without retraining.

Fig.~\ref{fig:thierarchy} shows samples generated from a varying the range of conditioning steps. Here, a narrower range allows image sampling from a broader subset following Eq.~\ref{prop:3}, resulting in diverse images. Conditioning in less than 500 steps, facial features differ from the references. 
The downsampling factor and conditioning range provide user controllability, where the later has a finer control on sample diversity.

\subsection{Multi-Domain Image Translation}\label{sec:domainshift}

Image-to-Image translation aims to learn the mapping between two visual domains. 
More specifically, generated images need to take the texture of the target domain while preserving the structure of the input images~\cite{park2020contrastive}. ILVR performs this task by matching the coarse information in reference images. We chose N=32 to preserve the coarse structure of the reference. 

The first two rows in Fig.~\ref{fig:domainshift} show samples generated with DDPM model trained on the FFHQ~\cite{stylegan} dataset, which contains high-quality photos of human faces. Samples from portrait~\cite{karras2020training} references show successful translation into photo-realistic faces. We also generated portraits from photos, with DDPM trained on METFACES~\cite{karras2020training}, the dataset of face portraits. Here, diverse samples are generated, however, some existing image translation models fail~\cite{isola2017image,park2020contrastive} to produce stochastic samples.

Generally, image translation models~\cite{isola2017image,zhu2017unpaired,liu2019few,choi2020stargan}, including multi-domain translation models~\cite{choi2020stargan,huang2018multimodal}, learn translation between different domains. Thus they can only translate from domains learned in the training phase. However, ILVR requires only a single model trained on the target domain. Therefore ILVR enables image translation from unseen source domains, with reference images from the low-resolution space of learned dataset as suggested in Eq.~\ref{prop:1}. Quantitative comparison to existing translation models is presented in the supplementary materials.

With a DDPM model trained on AFHQ-dog~\cite{choi2020stargan}, we translated images of dogs, cats, and wildlife animals from the validation set. The fourth to sixth row of Fig.~\ref{fig:domainshift} show the results. 
DDPM model trained only on dog images translates unseen cat and wildlife images well into dog images.

\begin{figure}[t!]
    \centering
    \includegraphics[width=1.0\linewidth]{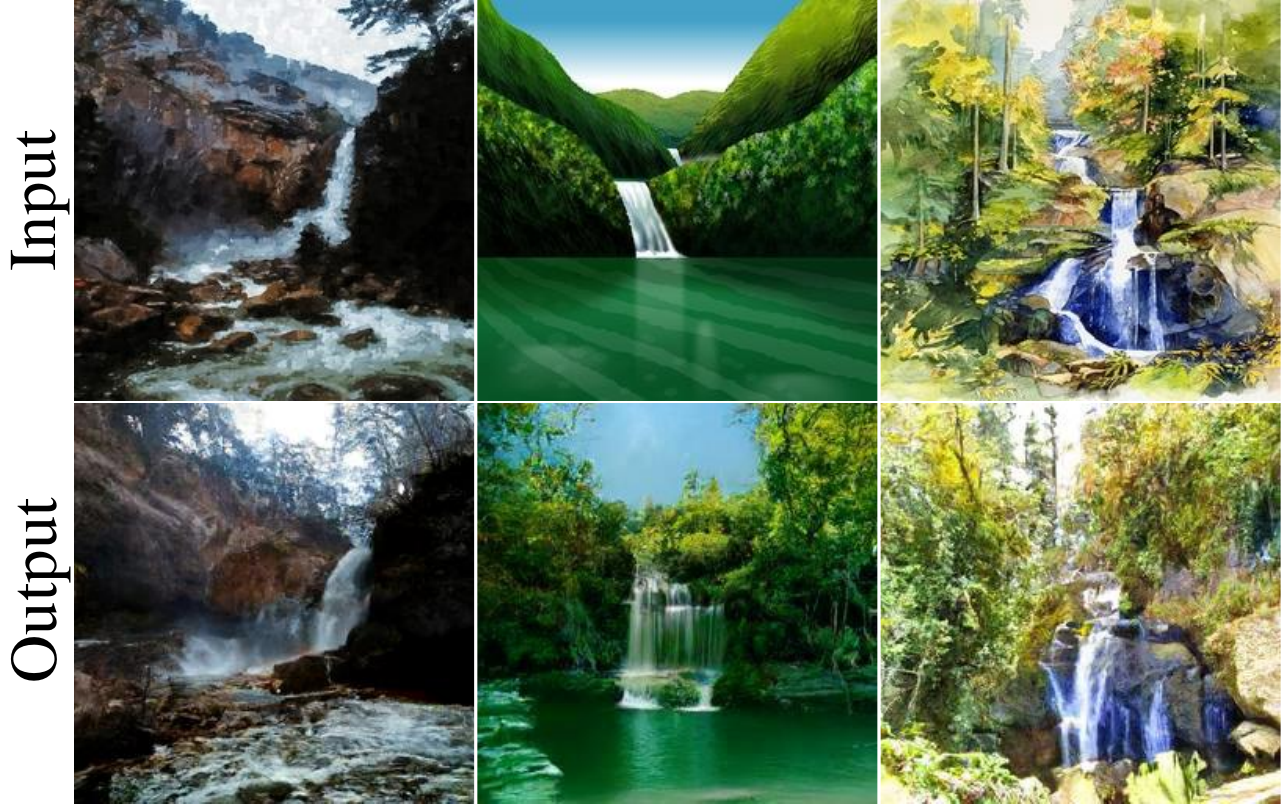}
    \caption{\textbf{Paint-to-Image.} Photo-realistic images generated from unnatural images (oil painting, water color, clip art)}
    \label{fig:paint2img}
    \vspace{-1em}
\end{figure}

\begin{figure}[t!]
    \centering
    \includegraphics[width=1.0\linewidth]{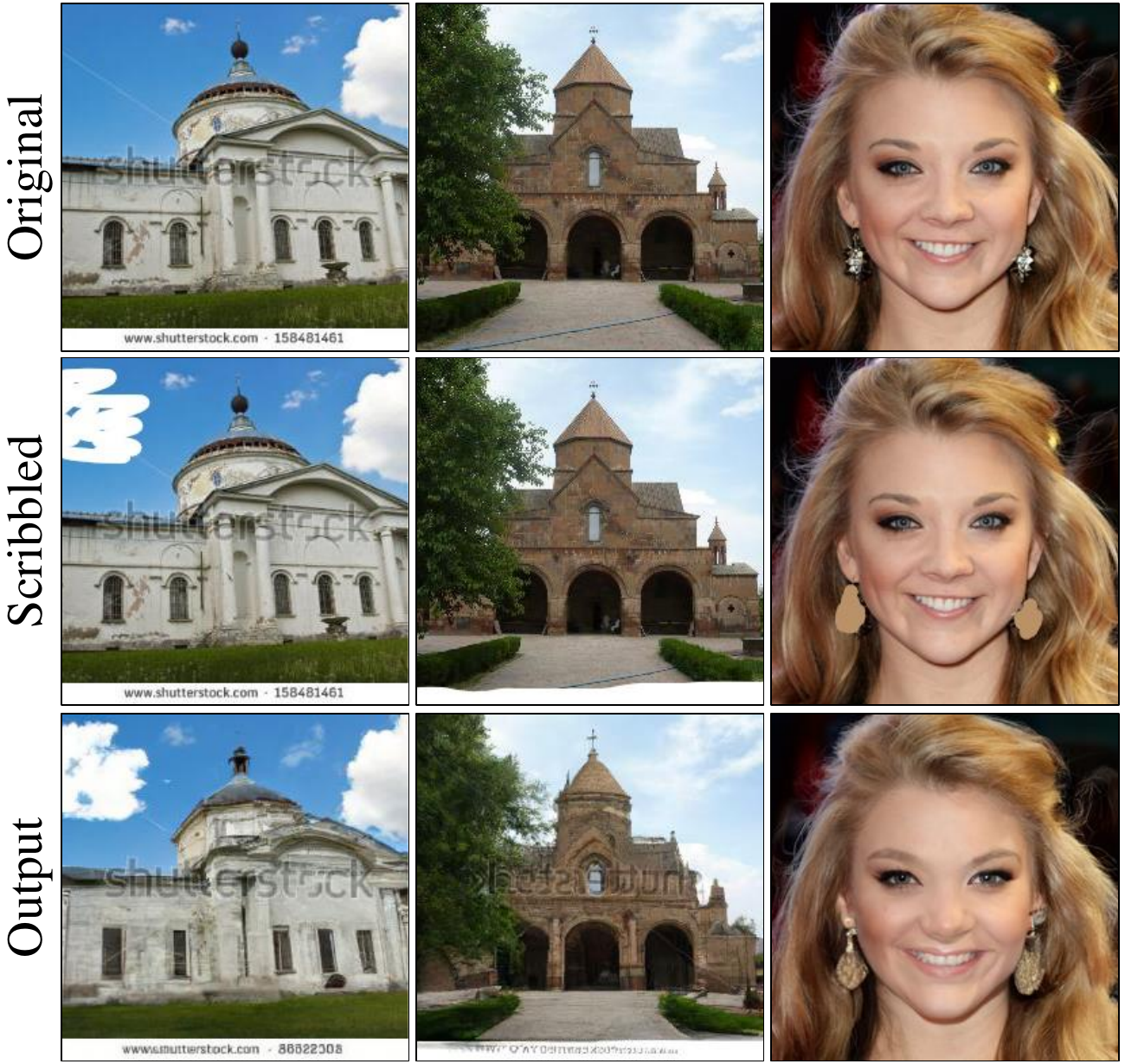}
    \caption{\textbf{Editing with scribbles.} Row1: cloud generated from white scribble at top-left corner. Row2: watermark generated from white scribble at the bottom. Row3: beige earring generated from scribble at top of silver earring.}
    \label{fig:scribble}
    \vspace{-0.5em}
\end{figure}

\subsection{Paint-to-Image}\label{sec:paint}
Paint-to-image is the task of transferring unnatural paintings into photo-realistic images. We validate our extension on this task using a model trained on the waterfall category from Places365~\cite{zhou2017places}.

As shown in Fig~\ref{fig:paint2img}, clip art, oil painting, and watercolor are well translated into photo-realistic images. 
Paintings and photo-realistic images differ in detailed texture. 
We chose a factor of N=64 to preserve only the coarse aspect (e.g., color scheme) of the reference. 
From Eq.~\ref{prop:1}, we can infer that the given paintings share coarse features of the learned dataset.

\subsection{Editing with Scribbles}\label{sec:scribble}

We extend our method to application of performing editions with user scribbles, which was also presented in Image2StyleGAN++ \cite{abdal2020image2stylegan++}. 
We generated samples with DDPM trained on LSUN-Church~\cite{yu2015lsun} and FFHQ~\cite{stylegan}. On reference images from the validation set, we added scribbles. Then, scribbled images are provided as references in factor N=8 on time steps from 1000 to 200, in order to both maintain details of original images and harmonize the scribbles. Interesting samples are shown in Fig.~\ref{fig:scribble}. In the second row, DDPM generated the "Shutterstock" watermark in the middle and the article number at the bottom. Since these pair of watermark and article number is common in the dataset, DDPM generated such features from a white scribble at the bottom. See supplementary for more samples.

\subsection{Quantitative Evaluation}\label{sec:evaluation}

We evaluated the quality and diversity of our generated images with widely used FID~\cite{heusel2017gans} and LPIPS~\cite{zhang2018unreasonable}.
The FID score evaluates the visual quality and distance between real and generated image distributions. LPIPS measures the perceptual similarity between two images.

Table~\ref{table:fid} reports FID scores measured from each downsampling factor $N$ and unconditional generation with models trained on FFHQ~\cite{stylegan} and METFACES~\cite{karras2020training} datasets. 
Scores (lower is better) are mostly comparable to the unconditional models, suggesting that our conditioning method does not harm the generation quality of unconditional model. In addition, FID scores of lower downsampling factors are better, as generated images from lower factors align almost perfectly with reference images. 

To evaluate the diversity among samples generated from the same
reference, we generated 10 images for each reference image and calculated average pairwise (45 pairs) LPIPS distance, following StarGAN2~\cite{choi2020stargan}. Table~\ref{table:lpips} shows that the higher the factor $N$, the higher the LPIPS, thus more diverse samples are generated as suggested in Eq.~\ref{prop:2}. In contrary, samples from lower $N$ share more amount of contents from the references, therefore less diverse.

\begin{table}[]
\centering
\begin{tabular}{|c|c|c|}
\hline
         & FFHQ & METFACES \\ \hline
baseline & 11.38 & 37.39        \\ \hline
4        & 4.62    & 11.85            \\
8        & 7.24  & 16.94              \\
16       & 10.35   & 22.68              \\
32       &  12.05   &  32.77           \\ \hline
\end{tabular}
\caption{FIDs of various downsampling factors and unconditional models. Models are trained on FFHQ and METFACES. Comparing to unconditional models (baseline), our conditioning maintains visual quality.}
\label{table:fid}
\end{table}
\begin{table}[!]
\centering
\begin{tabular}{|c|c|c|c|c|c|c|}
\hline
 N & 1 & 2 & 4 & 8 & 16 & 32 \\ \hline
 LPIPS & 0.011 & 0.039 & 0.101 & 0.185 & 0.299 & 0.439    \\ \hline
\end{tabular}
\caption{LPIPS distances computed on various downsampling factors. Higher $N$ results in higher diversity.}
\label{table:lpips}
\vspace{-1em}
\end{table}
\section{Related Work}

\subsection{Iterative generative models}

Successful iterative generative models gradually add noise to the data and learn to reverse this process. Score-based models~\cite{song2019generative, song2020improved} estimate a score (gradient of log-likelihood), and sample images with Langevin dynamics. A denoising score matching~\cite{vincent2011connection} is utilized to learn the scores in a scalable manner.
DDPM~\cite{sohl2015deep, ho2020denoising} learns to reverse the diffusion process that corrupts data, and utilizes the same functional form of the diffusion and reverse process. Ho~\etal~\cite{ho2020denoising} show superior performance in image generation, by achieving exceptionally low FID. Diffusion models also show superior performance in other domains such as speech synthesis~\cite{chen2020wavegrad,kong2020diffwave} and point cloud generation~\cite{luo2021diffusion}.
Our conditioning method allows this powerful DDPM to be utilized for a variety of purposes. 

\subsection{Conditional generative models}
Depending on the input type, such as class-label~\cite{van2016conditional, brock2018large, zhang2019self}, segmentation mask~\cite{wang2018high, park2019semantic}, feature from classifier~\cite{shocher2020semantic, nguyen2017plug}, and image~\cite{menon2020pulse, isola2017image}, various conditional generative models are available.
The studies employing images as a condition began with Isola~\etal~\cite{isola2017image}, and extended to unsupervised~\cite{zhu2017unpaired,huang2018multimodal}, few-shot~\cite{liu2019few}, and multi-domain image translations \cite{choi2020stargan}. Concurrent to our work, SR3~\cite{saharia2021image} trained conditional DDPM for super-resolution. These models show remarkable performances, however, only in the desired setting. In contrast, we demonstrate adaptation of a single unconditional model to various applications.

\subsection{Leveraging unconditional models}

Researches on leveraging pre-trained unconditional generators for various purposes, such as image editing~\cite{abdal2020image2stylegan++,shen2020interpreting}, style transfer~\cite{abdal2019image2stylegan}, and super-resolution~\cite{menon2020pulse,gu2020image} are being conducted. Specifically, by projecting given images into the latent vectors~\cite{zhu2020domain,abdal2019image2stylegan,abdal2020image2stylegan++,brock2016neural,yeh2017semantic} and manipulating them~\cite{harkonen2020ganspace,shen2020interpreting,menon2020pulse,gu2020image,choi2021toward}, images are easily edited. Leveraging capability of the unconditional models, these works exhibit high-quality images. GAN~\cite{goodfellow2014generative} forms a cornerstone of these works. However, we utilized the iterative generative model, DDPM, which has not been explored in this context.

\subsection{High-level semantics}

Image semantics contained in CNN features~\cite{shocher2020semantic}, segmentation masks~\cite{park2019semantic}, and low-resolution images~\cite{wang2020high,menon2020pulse} are actively used as conditions in generative models.
From our derivation in Eq.~\ref{eq:jointgivenc}, various kind of semantic conditions, such as features or segmentations, can provide high-level semantics. However, they require additional models (classifier or segmentation models). Since we are interested in controlling DDPM without any additional models, we provided semantics with a low-resolution image by utilizing iterative nature of DDPM.

\section{Conclusion}

We proposed a learning-free method of conditioning the generation process of unconditional DDPM. By refining each transition with given reference, we enable sampling from the space of plausible images. 
Further, downsampling factors and the conditioning range provide user controllability over this method. We demonstrated that a single unconditional DDPM can be leveraged to various tasks without any additional learning and models.
\vspace{0.9em}

\begin{figure*}[t!]
    \centering
    \includegraphics[width=1.0\linewidth]{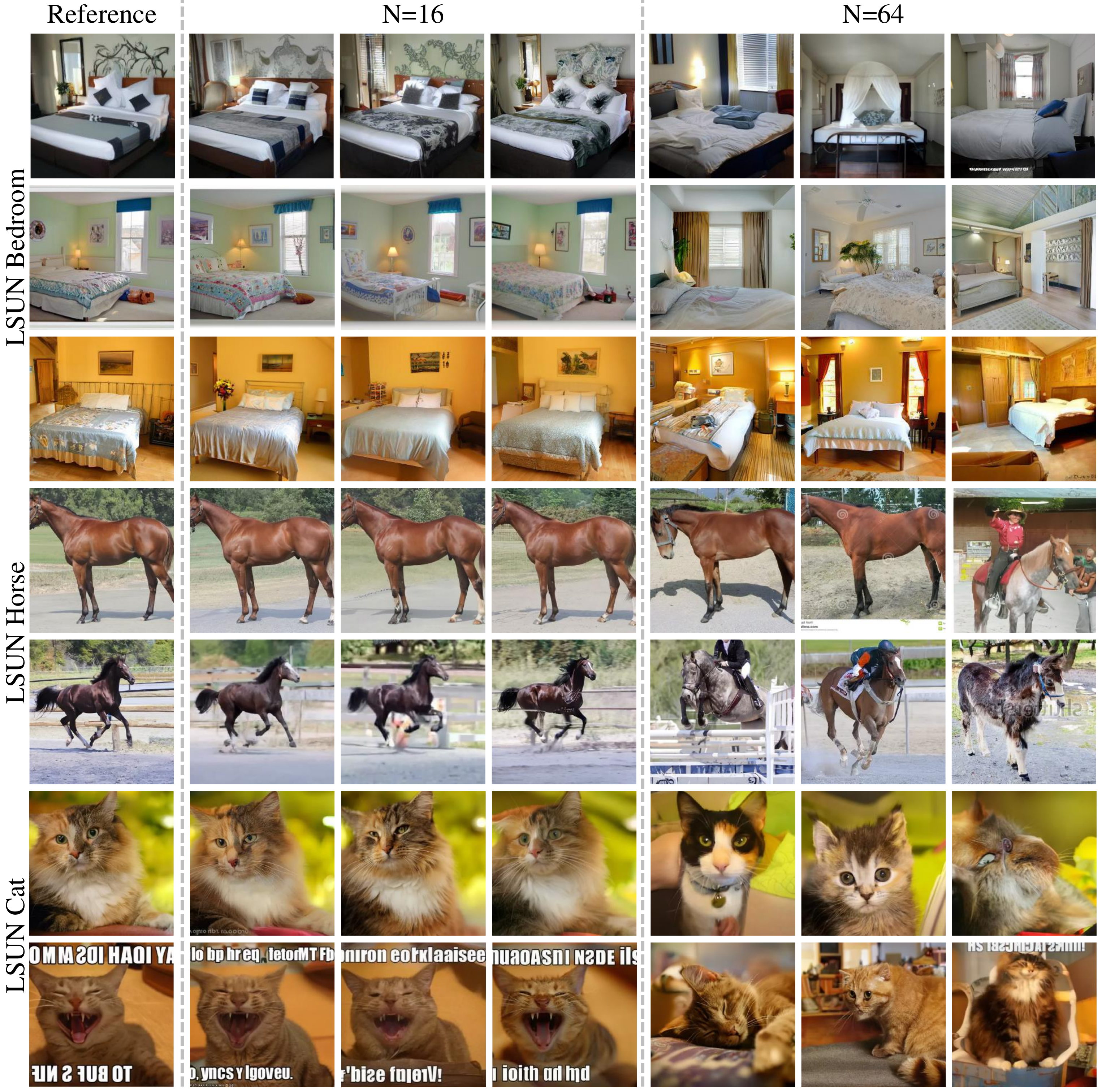}
    \caption{\textbf{ILVR samples with guided-diffusion~\cite{dhariwal2021diffusion}.} Publicly available guided-diffusion trained on LSUN Bedroom, Horse, and Cat datasets. For efficiency, samples are generated with 250 steps using uniform stride, following IDDPM~\cite{nichol2021improved}. Conditions are given in factor N=16,64 from time step 250 to 100. Samples share either coarse or fine semantics from the references.}
    \label{fig:gddpm}
    \vspace{7em}
\end{figure*}

\textbf{Acknowledgements:} This work was supported by Samsung Research Funding \& Incubation Center of Samsung Electronics under Project Number SRFC-IT1901-12, the National Research Foundation of Korea (NRF) grant funded by the Korea government (Ministry of Science and ICT)[2018R1A2B3001628], AIRS Company in Hyundai Motor and Kia through HMC/KIA-SNU AI Consortium Fund, and the BK21 FOUR program of the Education and Research Program for Future ICT Pioneers, Seoul National University in 2021.

\setcounter{section}{0}
\renewcommand\thesection{\Alph{section}}
\setcounter{table}{0}
\renewcommand{\thetable}{\Alph{table}}
\setcounter{figure}{0}
\renewcommand{\thefigure}{\Alph{figure}}
\setcounter{equation}{0}
\renewcommand{\theequation}{\Alph{equation}}

\section{Derivation of approximation}
In the main paper, we proposed iterative latent variable refinement (ILVR), where each transition of the generative process is matched with a given reference image.
Condition in each transition was replaced with a local condition based on our approximation, as suggested in Eq.7 of the main text.

Before detailed derivations of the approximation (Eq.7), we review notations used in the main text. With pre-defined hyperparameter $\overline{\alpha}_{t}$, latent variable $x_{t}$ can be sampled in closed-form: $x_{t}\sim q(x_{t}|x_{0})$ (Eq.2). Trained model $\epsilon _{\theta}(x_{t},t)$ predicts noise added in $x_{t}$, conditioned with time step $t$. 

From the property of the forward process that latent variable $x_{t}$ can be sampled from $x_{0}$ in closed-form, denoised data $x_{0}$ can be approximated with model prediction $\epsilon_{\theta}(x_{t},t)$:
\begin{align}\label{eq:s1}
x_{0}\approx f_{\theta}(x_{t},~t)=(x_{t}-\sqrt{1-\overline{\alpha}_{t}}~\epsilon_{\theta}(x_{t},t))/\sqrt{\overline{\alpha}_{t}}
\end{align}
Below is a derivation of Eq.7, where we approximated each conditioned Markov transition. We denote $\phi_{N}$ as $\phi$ and $f_{\theta}(x_{t},~t)$ as $f(x_{t})$ for brevity. From Eq.~\ref{eq:s1}, each conditional Markov transition with given reference image $y$ can be approximated as follows:
\begin{align}
&p_{\theta}(x_{t-1}|x_{t},\phi(x_{0})=\phi(y))\nonumber \\
\approx~&p_{\theta}(x_{t-1}|x_{t},\phi(f(x_{t-1}))=\phi(y))\nonumber \\
\approx~&\mathbb{E}_{q(y_{t-1}|y)}[p_{\theta}(x_{t-1}|x_{t},\phi(f(x_{t-1}))=\phi(f(y_{t-1})))].\nonumber
\end{align}
With linear property of operation $\phi$ and Eq.~\ref{eq:s1}, we have
\begin{align}
&\mathbb{E}_{q(y_{t-1}|y)}[p_{\theta}(x_{t-1}|x_{t},\phi(f(x_{t-1}))=\phi(f(y_{t-1})))]\nonumber \\
=~&\mathbb{E}_{q(y_{t-1}|y)}[p_{\theta}(x_{t-1}|x_{t},\phi(x_{t-1})=\phi(y_{t-1}),\nonumber \\
&\phi(\epsilon_{\theta}(x_{t-1}))=\phi(\epsilon_{\theta}(y_{t-1})))]\nonumber \\
\approx~&\mathbb{E}_{q(y_{t-1}|y)}[p_{\theta}(x_{t-1}|x_{t},\phi(x_{t-1})=\phi(y_{t-1}))].\nonumber
\end{align}
As shown in Eq.8 and Algorithm 1 of the main text, we first compute unconditional proposal $x_{t-1}'$, then refine it by ensuring $\phi(x_{t-1})=\phi(y_{t-1})$. Therefore,
\begin{align}
&\mathbb{E}_{q(y_{t-1}|y)}[p_{\theta}(x_{t-1}|x_{t},\phi(x_{t-1})=\phi(y_{t-1}))]\nonumber \\
=~&\mathbb{E}_{q(y_{t-1}|y)}[p_{\theta}(\phi(y_{t-1})+(I-\phi)(x_{t-1}') \nonumber\\
&|x_{t},\phi(x_{t-1})=\phi(y_{t-1}))]\nonumber \\
=~&\mathbb{E}_{q(y_{t-1}|y)}[p_{\theta}(x_{t-1}'|x_{t})]\nonumber \\
=~&p_{\theta}(x_{t-1}'|x_{t})\nonumber \\
=~&p_{\theta}(x_{t-1}|x_{t},\phi(x_{t-1})=\phi(y_{t-1})).\nonumber
\end{align}

\vspace{5em}
\section{Additional evaluations}
\begin{figure*}[t!]{}
    \centering
    \includegraphics[width=1.0\linewidth]{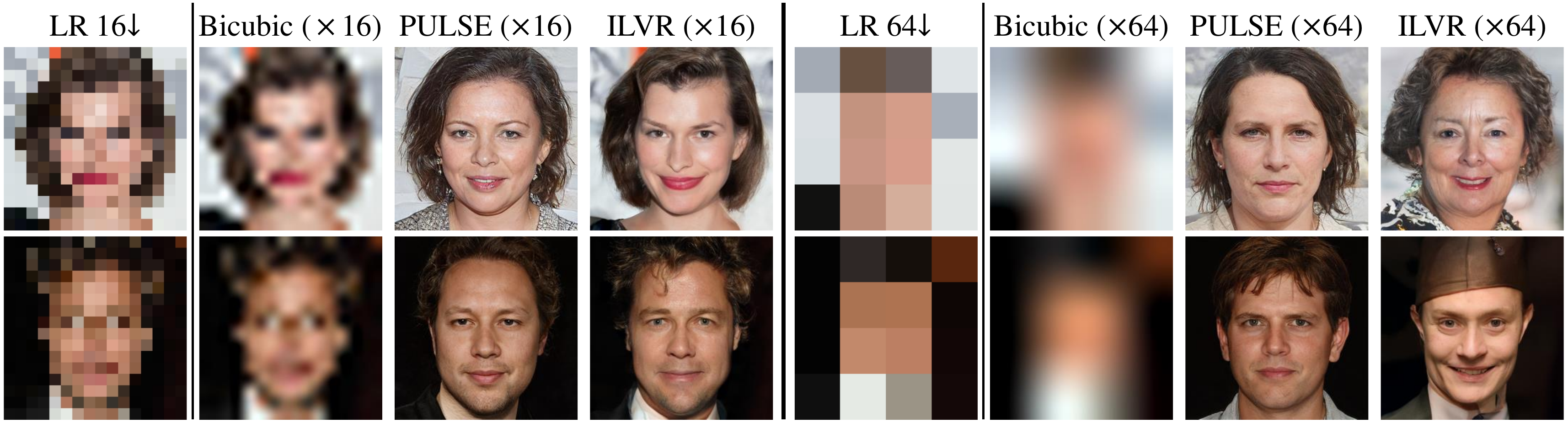}
    \caption{\textbf{Qualitative comparison on generation quality.} Images generated from reference images downsampled by a factor of 16 and 64. From LR images, ILVR generates faces with detailed features.}
    \vspace{-1em}
    \label{fig:pulse}
\end{figure*}

\subsection{Generation quality}
We provide additional qualitative and quantitative evaluations on the generation quality of ILVR. We evaluate images generated from low-resolution (LR) images downsampled by a factor of 16 and 64. Here, we compare ILVR with bicubic interpolation and PULSE~\cite{menon2020pulse}, a super-resolution study that leverages pre-trained StyleGAN~\cite{stylegan}. PULSE finds a latent vector that generates an image that downscales to the given LR image. We used publicly available StyleGAN2~\cite{karras2020analyzing} model\footnote{\url{https://github.com/rosinality/stylegan2-pytorch}} trained at $256\times 256$. Combining loss function from PULSE and StyleGAN2, we search for latent vectors with a loss as follows:
\begin{align}\label{eq:pulse_loss}
L_{total}=&||\phi(G(z))-\phi(y)||_2^2 \nonumber \\
&+ GEOCROSS(v_1,...,v_{14}) + \alpha L_{noise},
\end{align}
where each term refers to mean square error (MSE), geodesic cross loss~\cite{menon2020pulse}, and noise regularization~\cite{karras2020analyzing}, respectively. MSE ensures generated image $G(z)$ and reference image $y$ to match at low-resolution space. The geodesic cross loss ensures the latent vectors $v_1,...,v_{14}$ remain in the learned latent space. Noise regularization $L_{noise}$ discourages signal sneaking into the noise maps of StyleGAN2. We chose $\alpha = 5e^3$. Refer to StyleGAN2 literature for details on the noise regularization. We inherited initialization and learning rate schedule from StyleGAN2.

\begin{figure}[t!]{}
    \centering
    \includegraphics[width=1.0\linewidth]{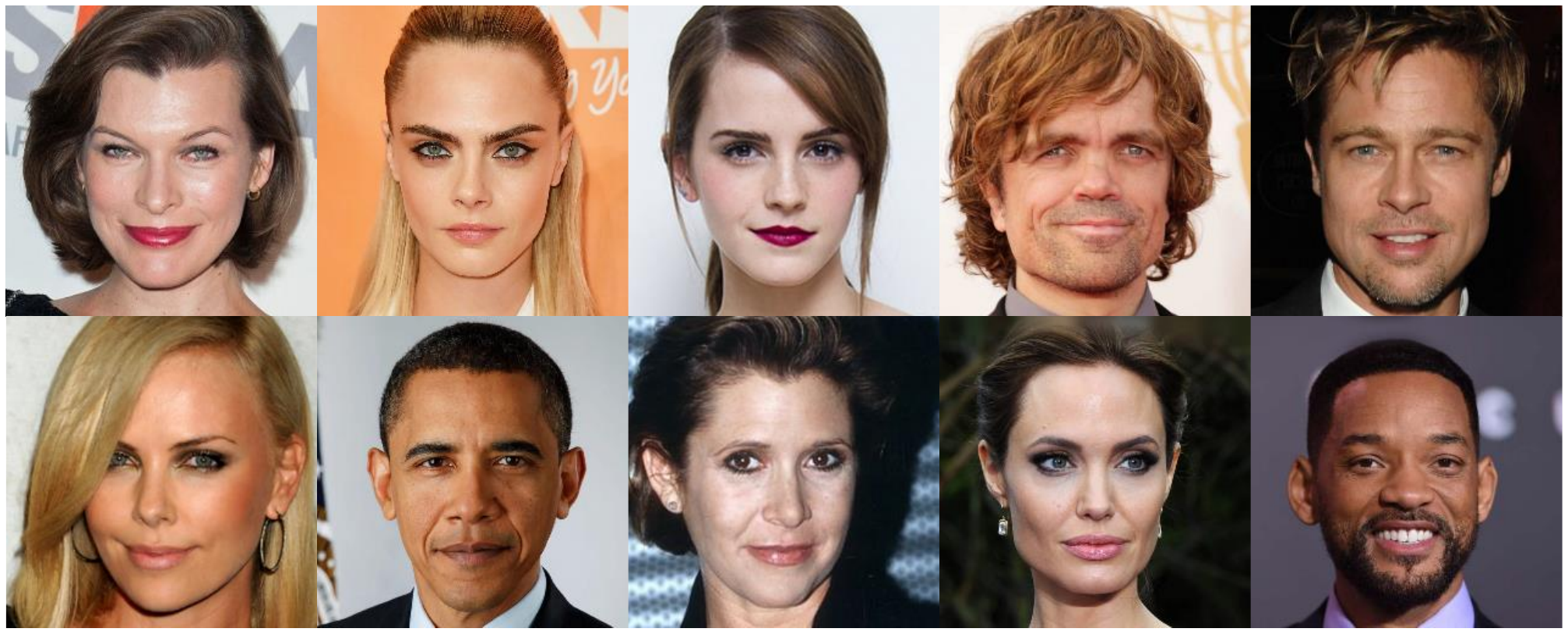}
    \caption{Images used for NIQE score.}
    \vspace{-0.5em}
    \label{fig:pulse-sample}
    
\end{figure}

Fig.~\ref{fig:pulse} presents additional qualitative results. ILVR and PULSE both show high-quality images generated from extremely downscaled images. Table.~\ref{table:niqe} shows NIQE~\cite{mittal2012making} score, which is a no-reference metric that measures the perceptual quality of an image. ILVR shows higher perceptual quality, even better than the original $256^2$ reference images (HR). We measured NIQE with reference images in Fig.~\ref{fig:pulse-sample}.

\begin{table}[]
\centering
\begin{tabular}{c|ccccc}
\hline
 $N$  & HR   & Nearest & Bicubic & PULSE & ILVR \\ \hline
$16\downarrow$ & 5.25 & 17.56   & 8.09    & 4.34  & \textbf{4.06}  \\
$64\downarrow$ & 5.25 & 14.15   & 12.45   & 4.10  & \textbf{4.02}  \\ \hline
\end{tabular}
\caption{\textbf{NIQE comparison on generation quality.} Lower is better. Scores measured with generated images from reference images downscaled by a factor of 16 and 64. ILVR exhibits the highest perceptual quality.}
\label{table:niqe}
\end{table}

\begin{table}[]
\centering
\begin{tabular}{|c|c|c|c|}
\hline
 CycleGAN~\cite{zhu2017unpaired} & MUNIT~\cite{huang2018multimodal}  & CUT~\cite{park2020contrastive}  & Ours \\ \hline
85.9     & 104.4 &  \textbf{76.2} & \textbf{79.8} \\ \hline
\end{tabular}
\caption{\textbf{FID comparison on image translation.} FID measured with images translated from test set of AFHQ-dog. ILVR is comparable to a state-of-the-art model.}
\vspace{-1em}
\label{table:i2i-fid}
\end{table}

\subsection{Image translation}
We compare Frechét inception distance (FID)~\cite{heusel2017gans} with image translation models on cat-to-dog (AFHQ~\cite{choi2020stargan} dataset) translation. Table.~\ref{table:i2i-fid} shows the results. FID scores are calculated with the test set from AFHQ~\cite{choi2020stargan}. ILVR presents comparable performance to CUT~\cite{park2020contrastive}, which is a state-of-the-art on cat-to-dog translation. Note that ILVR requires a model trained only on dog images, unlike the other models trained on both cat and dog images. We expect our result to broaden the applicability of DDPM to such image translation tasks.

\begin{figure*}[t!]{}
    \centering
    \includegraphics[width=1.0\linewidth]{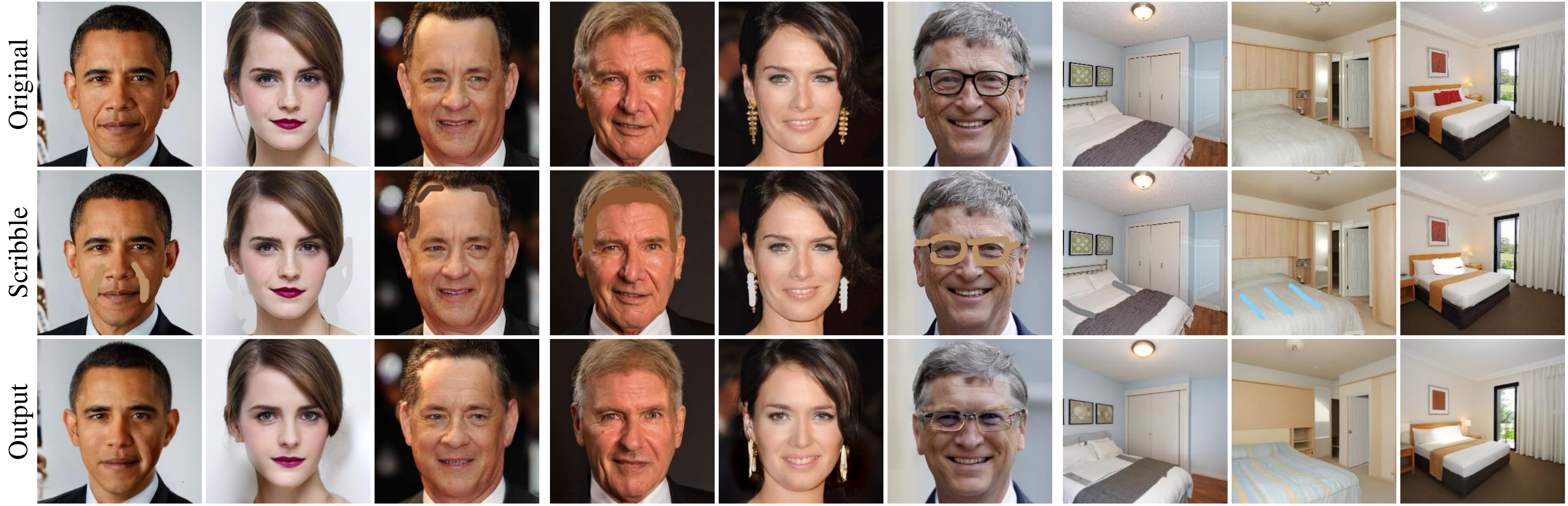}
    \caption{\textbf{Additional editing with scribbles.} Faces generated with our reproduced model trained on FFHQ~\cite{stylegan}. Bedrooms generated with publicly available model~\cite{dhariwal2021diffusion} trained on LSUN Bedroom~\cite{yu2015lsun}.}
    \label{fig:more_scribble}
    \vspace{3em}
\end{figure*}

\subsection{Additional samples}
Fig.~\ref{fig:gddpm} shows samples generated with publicly available guided-diffusion~\cite{dhariwal2021diffusion} trained on LSUN~\cite{yu2015lsun} datasets. We present additional editing with scribbles in Fig.~\ref{fig:more_scribble}.

\section{Implementation details}
We trained unconditional DDPM with publicly available PyTorch implementation.\footnote{\url{https://github.com/rosinality/denoising-diffusion-pytorch}}

\subsection{Low-pass filters}
\begin{figure*}[t!]{}
    \centering
    \includegraphics[width=0.95\linewidth]{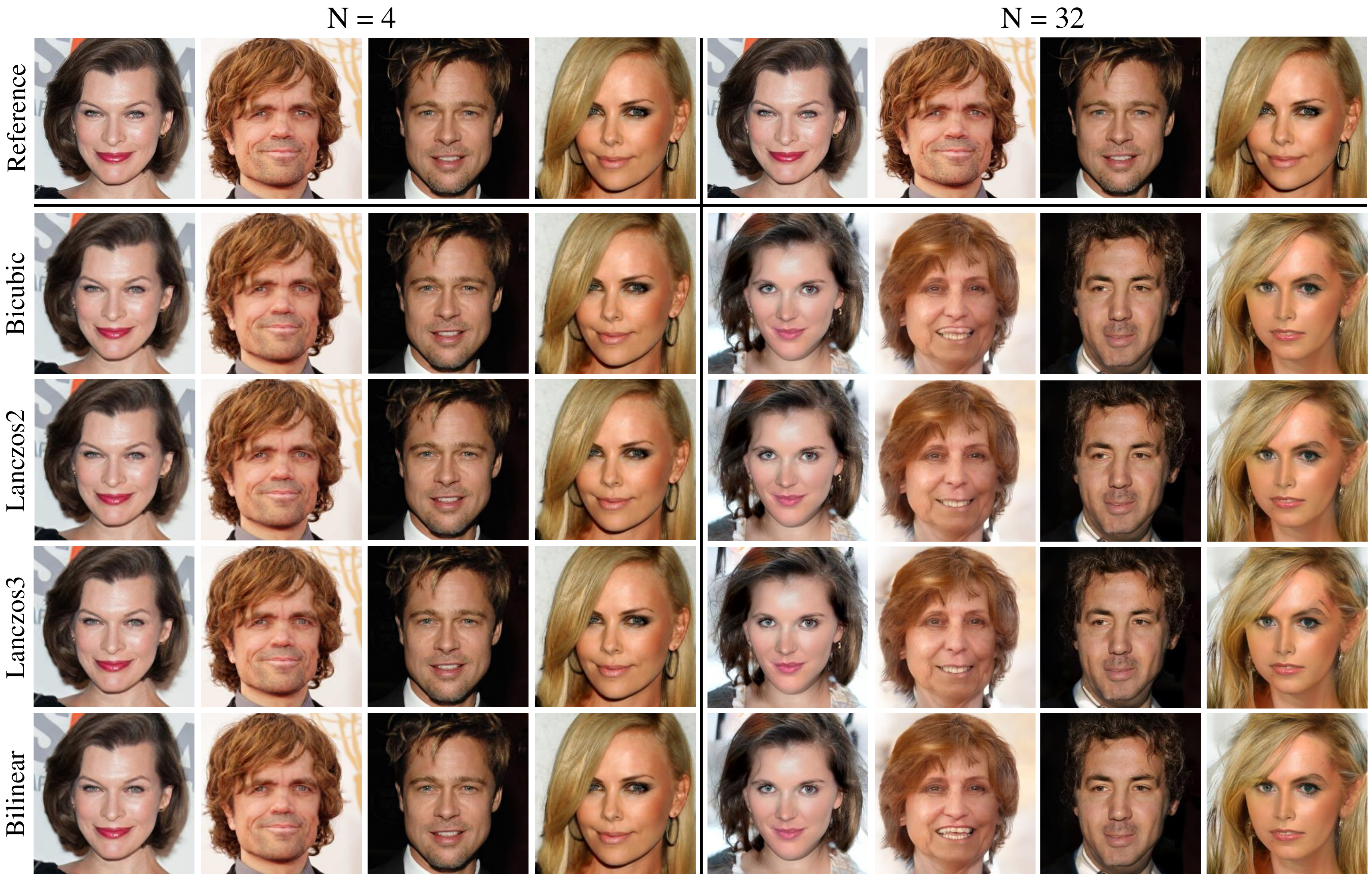}
    \caption{\textbf{Ablation on low-pass filters.} First column set: samples from downsampling factor N=4; Second column set: samples from downsampling factor N=32. Samples are generated with bicubic, lanczos2, lanczos3, bilinear interpolation for downsampling and upsampling. There is only a minor difference among filters, such as the exact position of teeth and hair.}
    \label{fig:lpfvar}
    \vspace{3em}
\end{figure*}

We used bicubic downsampling and upsampling with correctly implemented function~\cite{ResizeRight}. In Fig.~\ref{fig:lpfvar}, we compare generated samples where the same noises were added through the generative process, only differing resizing kernels. Among kernels, images are almost identical, suggesting that our method is robust to kernel choice.

\subsection{Datasets and training}
Here we describe datasets and training details. For all datasets, we trained at $256^2$ resolution with a batch size 8.

\textbf{FFHQ}~\cite{stylegan} consists of 70,000 high-resolution face images. We trained a model for 1.2M steps. 

\textbf{METFACES}~\cite{karras2020training} consists of 1,000 high-resolution portrait images. To avoid overfitting, we fine-tuned a model pre-trained on FFHQ~\cite{stylegan}, for 20k steps. 

\textbf{AFHQ}~\cite{choi2020stargan} consists of 15,000 high-resolution animal face images, which are equally split into three categories: dog, cat, and wild. We trained on the train set of dog category, then used test sets of three categories as reference images to demonstrate multi-domain image translation.

\textbf{Places365}~\cite{zhou2017places} consists of 10M images of over 400 scene categories. We trained a model on a waterfall category, which consists of 5,000 images. We used this model to paint-to-image task.

\textbf{LSUN Church}~\cite{yu2015lsun} consists of 126,227 images of churches. We trained a model for 1M steps.

\textbf{Paintings} used for paint-to-image task are collected from the web.

\subsection{Architecture}
We trained the same neural network architecture as Ho~\etal~\cite{ho2020denoising}, which is U-Net~\cite{ronneberger2015u} based on Wide ResNet~\cite{zagoruyko2016wide}. Details include group normalization~\cite{wu2018group}, self-attention blocks at $16\times 16$ resolution, sinusoidal positional embedding~\cite{vaswani2017attention}, and a fixed linear variance schedule $\beta_1,...,\beta_T$.

\subsection{Evaluation}
In Table 1 of the main text, we calculated FID with 50k real and 50k generated images using code\footnote{\url{https://github.com/mseitzer/pytorch-fid}} of PyTorch framework.


{\small
\bibliographystyle{ieee_fullname}
\bibliography{egbib}
}

\end{document}